\documentclass{preprint}

\usepackage{amsmath,amssymb,amsfonts}
\usepackage{graphicx}
\usepackage{booktabs}
\usepackage{listings}
\usepackage[utf8]{inputenc}
\usepackage[table]{xcolor}
\usepackage{mdframed}
\usepackage{colortbl}
\usepackage[ruled,vlined]{algorithm2e}
\usepackage{bbm}
\usepackage{pifont}
\usepackage[colorlinks=true]{hyperref}
\usepackage[numbers,comma,sort&compress]{natbib}

\newcounter{prompt}
\newenvironment{prompt}[1][]{
\refstepcounter{prompt}
\begin{mdframed}[
innertopmargin=0pt, 
innerbottommargin=0pt,
innerleftmargin=2pt, 
innerrightmargin=2pt,
frametitleaboveskip=2pt,
frametitlebelowskip=2pt,
frametitle={\textbf{Prompt~\theprompt:} #1},
frametitlefont=\footnotesize,
frametitlerule=true,
backgroundcolor=gray!10]%
}
{%
\end{mdframed}%
}

\newcounter{example}
\newenvironment{example}[1][]{
\refstepcounter{example}
\begin{mdframed}[
innertopmargin=2pt, 
innerbottommargin=2pt,
innerleftmargin=2pt, 
innerrightmargin=2pt,
frametitleaboveskip=2pt,
frametitlebelowskip=2pt,
splittopskip=0pt,
frametitle={\textbf{Ex~\theexample:} #1},
frametitlefont=\footnotesize,
frametitlerule=true,
backgroundcolor=gray!10]%
\setlength{\parindent}{0pt}%
\setlength{\parskip}{.5em}%
\footnotesize\itshape
}
{%
\par%
\end{mdframed}%
}

\lstdefinelanguage{prompt}{
    breaklines=true,
    breakindent=0pt,
    basicstyle=\footnotesize\ttfamily,
    columns=fullflexible,
    upquote=false,
}

\title{Comparing LLM and Fine-Tuned Model Performance on NVDRS Circumstance Extraction with Varying Prompt Complexity}

\author[1,2]{Geoffrey Martin}
\author[1]{Xuan Zhong Feng}
\author[1,2,*]{Yifan Peng}
\affil[1]{Department of Population Health Sciences, Weill Cornell Medicine, New York, USA}
\affil[2]{Systems Engineering, Cornell University, New York, USA}
\affil[*]{Corresponding author(s). Email(s): \url{yip4002@med.cornell.edu}}

\begin{document}

\maketitle

\begin{abstract}
Suicide is a leading cause of death in the United States, and understanding the circumstances that precede it requires extracting structured information from death investigation narratives. Many of these circumstances require semantic inference beyond simple keyword matching.
We develop a ``Complexity Score'' algorithm that analyzes coding manual structure to predict when detailed prompts with full coding guidelines improve over name-only prompts. We then construct a hybrid approach that selects prompt strategy per circumstance.
We evaluate large language models (LLMs) against fine-tuned RoBERTa on 25 inferentially complex circumstances from the National Violent Death Reporting System (NVDRS). We found that LLMs substantially outperform on low-prevalence circumstances where training data is insufficient.
We further demonstrate that our framework generalizes across frontier LLMs, with GPT-5.2, Gemini 2.5 Pro and Llama-3 70B showing consistent performance patterns. These findings support a hybrid architecture where LLMs handle rare, inferentially complex circumstances while fine-tuned models handle common ones.
\end{abstract}


\section{Introduction}

The National Violent Death Reporting System (NVDRS) is the only state-based surveillance system linking data from death certificates, coroner/medical examiner reports, and law enforcement reports for violent deaths in the United States \cite{cdc2022nvdrs}. Since 2003, trained abstractors have manually coded over 60 circumstance variables from free-text narratives (referred to as death investigation reports) for each case, creating a rich dataset for epidemiological research on suicide risk factors \cite{stone2018vital}. However, manual abstraction is time-intensive and subject to inter-rater variability \cite{wang2024natural}, and the CDC's constrained budget threatens the sustainability of this coding effort \cite{gostin2025assault}. Automated extraction of circumstances from death investigation reports could reduce abstractor burden while maintaining data quality.

Recent work has increasingly applied natural language processing (NLP) to extract Social Determinants of Health (SDOH) from clinical text \cite{lybarger2023advancements, lituiev2023automatic, chen2025extraction}. Ontological frameworks like SDoHO provide structured vocabularies for representing social factors \cite{dang2023sdoho}, while transformer-based models have achieved strong performance on SDOH extraction from electronic health records \cite{consoli2025sdoh, keloth2025social, peng2025enhancing}. More recently, large language models have shown particular promise. For example, Guevara et al. \cite{guevara2024large} demonstrated that GPT-4 achieves strong performance on extracting SDOH categories such as housing and employment from clinical notes. Consoli et al. \cite{consoli2025sdoh} demonstrated that LLMs capture nuanced social factors that rule-based systems miss. Further, Gabriel et al. \cite{gabriel2024development} validated LLM-based classifiers for clinical social needs screening, and Keloth et al. \cite{keloth2025social} showed that LLMs can extract SDOH across multiple institutions with minimal adaptation. Collectively, these studies show that LLMs can extract social factors from clinical narratives, but focus on relatively straightforward SDOH categories where the target concept is explicitly stated (e.g., ``patient is homeless'' $\rightarrow$ housing instability), leaving inferentially complex and low-prevalence circumstances underexplored.

For NVDRS specifically, Wang et al. \cite{wang2023nlp} fine-tuned BioBERT for NVDRS circumstance extraction, achieving strong performance on high-prevalence categories but poor results on rare circumstances and those requiring multi-step reasoning. Xu et al. \cite{xu2024analyzing} extended this work to analyze social factors across population groups. However, fine-tuned classifiers struggle with such circumstances because they learn surface patterns rather than the underlying inference rules \cite{mccoy2019right}. Supervised models need sufficient examples to learn each category's decision boundary, and many NVDRS circumstances have fewer than 1000 positive cases. 

Circumstances coded in NVDRS death investigation reports present a more challenging extraction task for NLP because many circumstances require semantic inference beyond surface pattern matching---the text describes a situation, but the abstractor must reason about what that situation implies under a specific context \cite{schoene2024automatically}. For example, the circumstance `Household Substance Abuse' is coded only if three criteria are met: (1) the victim is a child under 18, (2) the substance use must involve someone other than the victim, and (3) that person resides in the same household. Therefore, a model seeing ``alcohol bottles found in the home'' must infer that for adult victims, this reflects the victim's own use and does not meet the coding criteria of `Household Substance Abuse'. Similarly, ``Caregiver Burden'' requires role inference: a statement such as ``wife had terminal cancer'' implies the victim was providing care, not receiving it. 

To address these challenges, we hypothesize that LLMs, with their capacity for zero-shot reasoning \cite{kojima2022large}, outperform fine-tuned models on circumstances requiring deep semantic inference, particularly when training data is scarce. We propose a framework with three components: (1) we extract definitions, coding guidance and examples from the NVDRS coding manual to construct detailed prompts; (2) we develop a Complexity Score algorithm that analyzes the structure of coding manual examples to predict when detailed prompts are needed versus when simpler name-only prompts suffice; and (3) we construct a hybrid approach that uses the Complexity Score to select prompt strategy per circumstance, without requiring empirical tuning on each task.

We evaluate our proposed framework to extract 25 inferentially complex circumstances on the NVDRS. Our hybrid approach achieves macro F1 of 0.893 compared to fine-tuned RoBERTa's 0.800, with the largest gains on low-prevalence circumstances. We compare GPT-5.2, Gemini 2.5 Pro, and Llama-3 70B and observe consistent performance patterns across all three frontier LLMs, demonstrating that our framework is model-agnostic. Among circumstances predicted to require detailed guidance, complex prompts improve F1 by 9.5 percentage points over name-only prompts, while the Complexity Score algorithm achieves 87\% accuracy on non-tie circumstances in predicting optimal prompt strategy.

\section{Materials and Methods}

\subsection{Framework Overview}

Our framework uses the NVDRS coding manual to guide prompt construction and selection. For each circumstance, we first extract the definition, coding guidance, and positive/negative examples from the manual. We then compute a \textit{Complexity Score} based on the linguistic structure of the negative examples, which identifies circumstances where the LLM's default interpretation is likely to produce false positives without explicit guidance. If the Complexity Score exceeds a predefined threshold, we use a complex prompt containing full coding guidelines; otherwise, we use a simple prompt containing only the circumstance name. This hybrid approach automatically selects appropriate prompts in real time.

\subsection{Dataset and Circumstance Selection}

We utilized restricted-access death investigation reports from the NVDRS \cite{cdc2022nvdrs}. Each report combines the coroner/medical examiner (CME) and law enforcement (LE) narratives for each suicide case. Our dataset comprises 191,696 narratives.

To isolate circumstances that require semantic inference under NVDRS coding guidelines, we excluded those that can be extracted from keyword matching or strongly lexical cues (e.g., ``Mental Health Problem'' indicated by terms like `depression' or `schizophrenia'), as well as crisis-timing variables requiring temporal reasoning beyond our current scope.

Our final analysis focuses on 25 circumstances that require semantic inference. These circumstances were selected because they involve \textit{entity resolution} (determining who is affected by a mentioned condition), \textit{role inference} (understanding relationship directionality), or \textit{implicit reasoning} (recognizing that described situations imply unstated circumstances or applying non-obvious exclusion criteria).

\subsection{Prompting Strategy}

We developed two prompting approaches based on the NVDRS Coding Manual. We used an LLM (GPT-5.2 in this study) to extract structured information from the manual for each circumstance: the definition, detailed coding guidance specifying inclusion and exclusion criteria, and representative examples of cases that should and should not be coded. This information populates our complex prompt templates.

The \textbf{simple prompt} (Prompt 1) uses only the circumstance name, testing whether the LLM's world knowledge is sufficient without task-specific guidance.

\begin{prompt}[Simple prompt that uses only the circumstance name]
\label{lst:simple}
\begin{lstlisting}[language=prompt]
You are classifying death investigation narratives for the presence of specific circumstances.

Code "Yes" if the circumstance is mentioned, implied, or can be reasonably inferred from the narrative.
Code "No" if there is no mention or indication of the circumstance.

When in doubt, code "Yes".

---

Is there any mention of "{circumstance_name}" in this narrative?

---
NARRATIVE: {narrative}
---

EVIDENCE: [Quote relevant text, or "None found"]
FINAL CODING: [Yes or No]
\end{lstlisting}
\end{prompt}

The \textbf{complex prompt} (Prompt 2) includes the circumstance definition, detailed coding guidance with inclusion/exclusion criteria, and examples of positive and negative cases (\texttt{[...]} denotes truncation for space). This provides the model with the full context needed to apply NVDRS coding rules.

\begin{prompt}[Complex prompt with full coding guidelines]
\label{lst:complex}
\begin{lstlisting}[language=prompt]
[...]
Is there any mention of "{circumstance_name}" in this narrative?

DEFINITION: {nvdrs_definition}
CODING GUIDANCE: {detailed_guidance}
EXAMPLES - CODE "YES": {positive_examples}
EXAMPLES - CODE "NO": {negative_examples}

NARRATIVE: {narrative}
[...]
\end{lstlisting}
\end{prompt}

\subsection{Complexity Score}

We developed a heuristic algorithm to predict the appropriate prompt strategy (complex vs. simple) from the coding manual structure (Algorithm~\ref{alg:trickiness}). The algorithm scores each circumstance based on the linguistic features of its ``CODE NO" examples from the coding manual, identifying exclusions that describe scenarios consistent with the target category. High scores indicate that the LLM's default interpretation is likely to produce false positives, suggesting the need for explicit corrective guidance from the NVDRS coding guidelines.

\begin{algorithm}[t]
\caption{Complexity Score}
\label{alg:trickiness}
\KwIn{List of ``CODE NO'' examples from the coding manual}
\KwOut{Score predicting the complexity of circumstance extraction}
$score \gets 0$\;
$positive\_words \gets \{$`used', `had', `was', `moved', `argued', `problems', `history', `mentioned', `occurred', `abuse', `stressor'$\}$\;
\ForEach{$ex$ in ``CODE NO'' examples}{
    $has\_positive \gets$ any $positive\_words$ in $ex$\;
    $has\_but \gets$ ``but'' in $ex$\;
    \If{$has\_positive$ \textbf{and} $has\_but$}{
        $score \gets score + 3$
    }
    \ElseIf{$has\_positive$}{
        $score \gets score + 2$
    }
    \If{``use that'' \textbf{or} ``use other'' in $ex$}{
        $score \gets score + 1$
    }
    \If{$ex$ starts with ``No'' \textbf{and} length $< 5$ words}{
        $score \gets score - 1$
    }
}
\Return{$score$}\;
\end{algorithm}

The scoring rules quantify how likely each exclusion type is to cause false positives without explicit guidance. NO examples combining positive language with ``but'' receive +3 points (e.g., ``Argued constantly \textit{but} no specific incident") as they partially match the category but fail on a specific criterion, making them the hardest to classify correctly. Positive language alone receives +2 points (e.g., ``Victim had financial difficulties''), as something relevant that is mentioned but should not be coded. Category redirects receive +1 points when the manual explicitly directs coders to use a different circumstance (e.g., ``Code intimate partner conflicts under Intimate Partner Problem, not Other Relationship Problem"). These indicate boundary cases where similar circumstances overlap. Simple absences receive -1 point (e.g., ``No financial difficulties.''). Unlike complex exclusions, these state directly that the circumstance is absent, requiring no inference; a coding manual dominated by such examples indicates straightforward classification. 

In this study, we set the decision threshold at 2, where circumstances scoring above 2 use complex prompts, while those at or below 2 use simple (name-only) prompts.

\subsection{Experimental Settings}

NVDRS is highly imbalanced, with some circumstances having fewer than 100 positive cases among 191,696 narratives. To enable fair evaluation of both precision and recall, we used balanced sampling. For each circumstance, we uniformly sampled 200 narratives (100 positive, 100 negative) for evaluation. All models were evaluated on the same sampled set to enable direct comparison. We report precision, recall, and F1 score with 95\% Wilson confidence intervals \cite{lam2024confidence}.

We compared five models. \textbf{RoBERTa} was fine-tuned as a multi-label classifier on 191,696 narratives with binary cross-entropy loss and class weights inversely proportional to label frequency. Training used a learning rate of $10^{-5}$, batch size of 12, max sequence length of 512, early stopping with patience of 3 epochs, and up to 20 training epochs. All LLM evaluations used temperature 0.3 with narratives truncated to 3500 characters. The test set (200 samples per circumstance, 100 positive and 100 negative) was held out before training and used identically for all models.
\textbf{GPT-5.2} was evaluated in zero-shot prompting with both complex and simple prompts.

To test the generalizability of our prompting framework, we also evaluated \textbf{Gemini 2.5 Pro} and \textbf{Llama-3 70B} with complex prompting. 

\section{Results and Discussion}

\begin{table}[t]
\centering
\caption{Macro F1 scores across approaches. Hybrid uses the Complexity Score algorithm to select complex or simple prompts per circumstance. Oracle selects the better prompt per circumstance with hindsight.}
\label{tab:overall}
\footnotesize
\begin{tabular}{llr}
\toprule
\textbf{Approach} & \textbf{Model} & \textbf{Macro F1} \\
\midrule
\rowcolor{gray!10} Oracle & GPT-5.2 & .897 \\
Training-based & RoBERTa & .800 \\
\rowcolor{gray!10} Simple Prompt & GPT-5.2 & .855 \\
Complex Prompt & Llama-3 70B & .838\\
& Gemini 2.5 Pro & .878\\
& GPT-5.2 & .883 \\
\rowcolor{gray!10}Hybrid Prompt & GPT-5.2 & \textbf{.893} \\
\bottomrule
\end{tabular}
\end{table}

\begin{table}[t]
\centering
\caption{F1 scores (95\% CI) for all model configurations across 25 circumstances, sorted by number of positive training instances of RoBERTa. $^*$--RoBERTa not evaluable for Household Substance Abuse due to insufficient held-out data.}
\label{tab:llm_results}
\scriptsize
\begin{tabular}{lrccccc}
\toprule
 & \multicolumn{2}{c}{\textbf{Training-based}} & \textbf{Simple} & \multicolumn{3}{c}{\textbf{Complex}} \\
\cmidrule(r){2-3}\cmidrule(r){4-4}\cmidrule{5-7}
\textbf{Circumstance} & \textbf{Training} & \textbf{RoBERTa} & \textbf{GPT-5.2} & \textbf{GPT-5.2 } & \textbf{Gemini 2.5 Pro} & \textbf{Llama-3 70B} \\
\midrule
Household Substance Abuse & 18 & --$^*$ & .309 (.210--.429) & \textbf{.812} (.630--.917) & .812 (.630--.917) & .762 (.597--.874) \\
Abuse or Neglect & 182 & .131 & .882 (.828--.920) & \textbf{.894} (.842--.931) & .850 (.793--.893) & .831 (.773--.877) \\
Caregiver Burden & 322 & .496 & \textbf{.954} (.914--.976) & .901 (.847--.937) & .897 (.844--.933) & .840 (.784--.884) \\
Living Situation Change & 402 & .481 & .752 (.685--.809) & \textbf{.845} (.786--.890) & .825 (.767--.871) & .795 (.734--.845) \\
Victim of Violence & 777 & .798 & .858 (.803--.900) & .817 (.757--.865) & \textbf{.924} (.878--.953) & .840 (.784--.884) \\
Traumatic Anniversary & 1,264 & .837 & \textbf{.948} (.906--.972) & .886 (.829--.926) & .931 (.885--.959) & .911 (.863--.943) \\
Other Addiction & 1,368 & .726 & \textbf{.784} (.721--.837) & .675 (.585--.754) & .699 (.614--.773) & .736 (.656--.802) \\
Physical Fight & 1,616 & .885 & \textbf{.956} (.917--.977) & .950 (.909--.973) & .919 (.873--.950) & .908 (.860--.941) \\
Family Stressor & 1,636 & .587 & .724 (.660--.780) & .776 (.711--.829) & \textbf{.787} (.726--.838) & .702 (.638--.758) \\
Disaster Exposure & 1,787 & .890 & .830 (.761--.883) & .942 (.898--.968) & \textbf{.953} (.913--.976) & .945 (.902--.969) \\
Childhood Abuse History & 2,122 & .895 & .964 (.927--.983) & .953 (.913--.976) & \textbf{.980} (.948--.992) & .919 (.871--.949) \\
Treatment Non-Adherence & 2,316 & .692 & .879 (.824--.919) & .880 (.824--.920) & \textbf{.885} (.832--.922) & .770 (.708--.822) \\
School Problem & 2,625 & .948 & .954 (.913--.976) & .959 (.921--.979) & \textbf{.970} (.935--.986) & .950 (.909--.973) \\
Other Relationship Problem & 3,878 & .776 & .767 (.704--.821) & \textbf{.836} (.773--.884) & .796 (.731--.849) & .790 (.727--.841) \\
Suicide of Friend/Family & 4,189 & .936 & \textbf{.970} (.935--.986) & .947 (.904--.972) & .964 (.927--.983) & .911 (.863--.943) \\
Civil Legal Problem & 5,921 & .875 & \textbf{.890} (.838--.926) & .839 (.776--.887) & .853 (.793--.898) & .858 (.802--.901) \\
Eviction or Housing Loss & 6,198 & .948 & \textbf{.965} (.929--.983) & .959 (.920--.979) & .947 (.907--.971) & .907 (.859--.940) \\
Death of Friend/Family & 11,104 & .944 & .952 (.913--.974) & \textbf{.971} (.937--.987) & .971 (.937--.987) & .853 (.798--.896) \\
Criminal Legal Problem & 13,951 & .910 & \textbf{.919} (.871--.950) & .913 (.864--.945) & .890 (.838--.926) & .856 (.800--.898) \\
Financial Problem & 14,597 & .913 & .887 (.834--.924) & \textbf{.913} (.864--.945) & .876 (.822--.915) & .877 (.823--.916) \\
Family Relationship Problem & 15,726 & .884 & .797 (.736--.847) & \textbf{.858} (.802--.901) & .824 (.765--.870) & .697 (.633--.754) \\
Job Problem & 16,813 & .943 & .970 (.936--.986) & \textbf{.980} (.950--.992) & .930 (.885--.958) & .921 (.875--.951) \\
Argument & 28,438 & .949 & .903 (.854--.937) & \textbf{.960} (.921--.980) & .900 (.851--.934) & .883 (.831--.921) \\
Physical Health Problem & 37,582 & \textbf{.925} & .792 (.732--.842) & .800 (.740--.849) & .820 (.761--.867) & .750 (.687--.804) \\
Depressed Mood & 61,249 & \textbf{.839} & .762 (.699--.815) & .797 (.735--.847) & .748 (.685--.802) & .742 (.679--.796) \\
\midrule
\textbf{Macro F1} & & .800 & .855 & \textbf{.883} & .878 & .838 \\
\bottomrule
\end{tabular}
\vspace{-1em}
\end{table}

\subsection{Main Results}

We evaluated five model configurations on 25 inferentially complex NVDRS circumstances to assess whether LLMs can outperform fine-tuned models on tasks requiring high semantic inference, and whether our Complexity Score algorithm can effectively select a prompt strategy.

Table~\ref{tab:overall} summarizes overall performance. The hybrid approach, which uses the Complexity Score algorithm to select a prompt strategy per circumstance, achieves the highest macro F1 (0.893). This closely approaches the oracle upper bound of 0.897, where the oracle is defined as selecting the better of two prompt strategies per circumstance with hindsight. 

Among single-strategy baselines, GPT-5.2 with complex prompting alone achieves the best macro F1 (0.883), followed by Gemini (0.878), GPT-5.2 with simple prompting (0.855), Llama-3 (0.838), and RoBERTa (0.800). The consistent performance across GPT-5.2, Gemini, and Llama-3 demonstrates that our framework generalizes across frontier LLMs.

\subsection{Complex Prompt Analysis}

Overall, complex prompting improves GPT-5.2 macro F1 by 2.8 points (0.883 vs. 0.855), but the effect is heterogeneous across circumstances. The largest improvement occurs for Household Substance Abuse, where complex prompting provides a +50.3 point improvement (Table~\ref{tab:llm_results}). The coding rules of this circumstance are counterintuitive: requiring a child victim, substance use by others, and the same household. Without explicit guidance, GPT-5.2 tends to code any mention of substance use as \textit{positive}. Similarly, improvements are also observed for Disaster Exposure (+11.2 points) and Living Situation Change (+9.3 points), both of which benefit from detailed definitions that clarify their non-obvious scope.

Conversely, some circumstances perform better with simple (name-only) prompts (Table~\ref{tab:llm_results}). In particular, Other Addiction (-10.9 points), Traumatic Anniversary (-6.2 points), and Caregiver Burden (-5.3 points) all show decrements with complex prompting. A likely explanation is that specific examples included in the complex prompt are overly restrictive, causing LLMs to generate false negatives for valid cases that do not match the examples exactly.

\subsection{Prompt Strategy Analysis}

We then evaluate the accuracy of the Complexity Score algorithm. Table~\ref{tab:complexity} shows that our algorithm achieves 72\% accuracy (18/25 circumstances). When excluding ties (circumstances where the F1 difference between complex and simple prompts is $\leq 0.02$), accuracy rises to 87\% (13/15 circumstances).

Among the 10 circumstances the algorithm identifies as needing complex prompts, using complex prompts yields an average F1 of 0.887, compared with 0.792 with simple prompting (+9.5 points). Conversely, among the 15 circumstances predicted to favor simple prompting (Table~\ref{tab:complexity}), simple prompts yield 0.897, outperforming 0.880 with complex prompts (+1.7 points). As a result, the hybrid approach achieves an overall F1 of 0.893, capturing most of the oracle's gain (0.897) over always using complex prompts (F1 0.883). These results demonstrate that prompt strategy can be selected algorithmically without empirical tuning for each circumstance.

\begin{table}[t]
\centering
\small
\caption{Complexity Score analysis. Score $>$ 2 predicts complex prompt (C), otherwise simple (S). Oracle indicates which prompt actually performed better.}
\newcommand{\xmark}{\textcolor{red}{\ding{55}}}%
\newcommand{\cmark}{\textcolor{teal}{\ding{51}}}
\label{tab:complexity}
\begin{tabular}{lrccc}
\toprule
\textbf{Circumstance} & \textbf{Score} & \textbf{Oracle} & \textbf{Pred} & \textbf{Result} \\
\midrule
Argument & 5 & C & C & \cmark \\
Depressed Mood & 5 & C & C & \cmark \\
Living Situation Change & 4 & C & C & \cmark \\
Disaster Exposure & 4 & C & C & \cmark \\
Other Relationship Problem & 4 & C & C & \cmark \\
Death of Friend/Family & 4 & C & C & \cmark \\
Family Relationship Problem & 4 & C & C & \cmark \\
Household Substance Abuse & 3 & C & C & \cmark \\
Abuse or Neglect & 3 & C & C & \cmark \\
Childhood Abuse History & 3 & S & C & \xmark \\
Other Addiction & 2 & S & S & \cmark \\
Family Stressor & 2 & C & S & \xmark \\
Suicide of Friend/Family & 2 & S & S & \cmark \\
Criminal Legal Problem & 2 & S & S & \cmark \\
Caregiver Burden & 1 & S & S & \cmark \\
Victim of Violence & 1 & S & S & \cmark \\
Civil Legal Problem & 1 & S & S & \cmark \\
Physical Health Problem & 1 & C & S & \xmark \\
Treatment Non-Adherence & 0 & C & S & \xmark \\
School Problem & 0 & C & S & \xmark \\
Job Problem & 0 & C & S & \xmark \\
Traumatic Anniversary & $-$1 & S & S & \cmark \\
Physical Fight & $-$1 & S & S & \cmark \\
Eviction or Housing Loss & $-$1 & S & S & \cmark \\
Financial Problem & $-$1 & C & S & \xmark \\
\midrule
\textbf{Accuracy} & & & & 18/25 \\
\bottomrule
\end{tabular}
\end{table}

\subsection{Training Sample Threshold}

\begin{table}[t]
\centering
\small
\caption{Hybrid vs RoBERTa by training set size bracket based on F1 point estimates. Household Substance Abuse counted as Hybrid win (RoBERTa not evaluable).
}
\label{tab:prevalence}
\begin{tabular}{lrrr}
\toprule
\textbf{Training Instances} & \textbf{n} & \textbf{Hybrid} & \textbf{RoBERTa} \\
\midrule
$<$ 500 & 4 & 4 & 0 \\
500--2,000 & 6 & 6 & 0 \\
2,000--5,000 & 5 & 5 & 0 \\
5,000--15,000 & 5 & 4 & 1 \\
$>$ 15,000 & 5 & 2 & 3 \\
\midrule
\textbf{Total} & 25 & 21 & 4 \\
\bottomrule
\end{tabular}
\end{table}

To understand how prevalence affects model performance, we analyzed win rates across training set size brackets. We hypothesized that LLMs outperform on low-prevalence circumstances where RoBERTa lacks sufficient training examples to learn decision boundaries.

Table~\ref{tab:prevalence} breaks down performance by circumstance prevalence. For circumstances with fewer than 5,000 positive training instances, the hybrid approach achieves a higher F1 than RoBERTa on all 15 circumstances. For circumstances with more than 5,000 training instances, the models are more competitive as RoBERTa wins 4 of 10 comparisons. These results suggest that, when training data is scarce, LLM-based approaches dominate, while fine-tuned models close the gap when sufficient training instances are available.

The largest hybrid approach advantages occur at the extreme low-prevalence end. For Abuse or Neglect (182 positive training instances), the hybrid approach improves F1 by 0.76, for Caregiver Burden (322 instances), by 0.46, and for Living Situation Change (402 training instances), by 0.36. For Household Substance Abuse, which has only 18 training instances, RoBERTa could not be evaluated due to insufficient held-out data, but the LLM achieves an F1 of 0.812 in a zero-shot setting. These results highlight  circumstances where training-based approaches struggle but LLM-based approaches provide substantial advantages.

RoBERTa's wins concentrate when there is an abundance of training instances. For example, it outperforms in Physical Health Problem (+0.13 F1, 37,582 training instances), Depressed Mood (+0.04 F1, 61,249 training instances), and Family Relationship Problem (+0.03 F1, 15,726 training instances), and Financial Problem (+0.03 F1, 14,597 training instances).

\subsection{Error Analysis}


Manual review of cases where model predictions disagreed with ground truth labels revealed three types of error: ground truth error, model error, and ambiguity.

\paragraph{Ground Truth Error}

In some cases, ground truth labels are categorically incorrect. Example \ref{exp:child} describes an adult financial conflict between the victim and her mother, including threats of eviction from a family-owned condominium. The statement ``I'm not doing homeless again'' may have led the coder to infer a difficult childhood, but the narrative provides no explicit evidence of childhood abuse or neglect, only housing instability during adulthood. Example \ref{exp:eviction} illustrates a potential data entry error, as the narrative contains no explicit or implied mention of housing instability, loss, or eviction. In both cases, all five models correctly identified the absence of these circumstances, suggesting that unanimous model disagreement with ground truth labels can serve as a useful initial signal for quality control in human-coded datasets.

\begin{example}[Childhood Abuse History (Label: Positive, All Models: Negative)]
\label{exp:child}
The mother reported they had been in an argument prior to the incident concerning finances. The decedent was unable to pay the association fees of the condominium she was residing in and the mother threatened to evict her. [...] The V sent an email to her mother the night before the incident stating `I'm not doing homeless again and I'm not living in fear of when you're going to sell the condo and put me out.'
\end{example}

\begin{example}[Eviction or Housing Loss (Label: Positive, All Models: Negative)]
\label{exp:eviction}
It was reported that V had lost his girlfriend and she filed a restraining order against him. At some point in the past, V violated the restraining order and was placed in jail for domestic violence. [...] The V was upset over losing his girlfriend and being arrested for domestic violence in the past.
\end{example}

\paragraph{Model Error}

In this category, models over-generalized from surface lexical patterns. Example \ref{exp:argument} describes a narrative where the victim's boyfriend ``confronted'' her about seeking addiction treatment. While ``confronted'' typically carries adversarial connotation, the narrative does not explicitly describe or imply a hostile dispute, i.e., an argument. Instead, the narrative suggests a potentially supportive intervention (``they planned to speak to her employer for assistance''). This suggests that models overemphasized the negative connotation associated with the word ``confronted'' while downplaying contextual clues. 
A similar pattern appears in Example \ref{exp:eviction2}, where the victim is described as having been ``kicked out''. However, contextual clues clarify that the couple had been ``living together'', suggesting a temporary dispute rather than actual eviction or housing loss. Together, these cases reveal a tendency by LLMs, even when given additional NVDRS guidance, to over-rely on lexical cues without recognizing when context invalidates or refines them.

\begin{example}[Argument (Label: Negative, All Models: Positive)]
\label{exp:argument}
V's boyfriend had confronted V about getting help with her addiction, and they planned to speak to her employer for assistance. Letters were found in V's bedroom dating back two years ago, stating how much pain V was in and saying goodbye to her loved ones.
\end{example}

\begin{example}[Eviction or Housing Loss (Label: Negative, All Models: Positive)]
\label{exp:eviction2}
Victim and victims girlfriend (witness) were living together. It was reported victim and witness had an argument the night prior, witness kicked victim out afterwards. [...] The victim was found the next day by neighbors near the top of the first set of back porch steps of his residence.
\end{example}

\paragraph{True Ambiguous Cases}

Other narratives contain information that could reasonably support more than one coding decision. Example \ref{exp:caregiver} describes a young father expressing guilt about ``not being there'' for his child. The phrasing is ambiguous as it could indicate feeling overwhelmed by parenting responsibilities or regret about his absence. The appropriate coding decision depends on whether the victim perceived parenting as a burden or a responsibility he failed to meet, which is difficult to infer from the narrative alone. In such cases, where the ground truth itself requires subjective judgment, model disagreement may reflect irreducible ambiguity rather than inadequate understanding.

\begin{example}[Caregiver Burden (Label: Positive, Models: Mixed)]
\label{exp:caregiver}
V's child's mother came to the scene and reported that they had been texting off and on before the incident. V had been apologizing for not being a good father and for not being there for her and the baby.
\end{example}

\subsection{Limitations}

Several limitations warrant caution. First, our evaluation uses 200 samples per circumstance due to API and time costs. Second, while Wilson confidence intervals suggest estimates are reliable ($\pm$0.04--0.06), larger samples would tighten these bounds. 
Finally, we evaluate on NVDRS data only; generalizability to other coding schemes remains an open question.

\section{Conclusion}

We develop a ``Complexity Score'' algorithm that analyzes coding manual structure to predict when detailed prompts improve over name-only prompts. We show that LLMs, with their capacity for zero-shot reasoning, outperform fine-tuned models on circumstances requiring deep semantic inference, particularly when training data is scarce. 
These findings support deploying LLMs for rare, inferentially complex circumstances while retaining fine-tuned models for common ones.

\section*{Acknowledgment}

Research reported in this work was partially funded through a Patient-Centered Outcomes Research Institute (PCORI) Award (ME-2023C3-35934), National Library of Medicine grant (R01LM014306, R01LM014573), and National Science Foundation Graduate Research Fellowship under Grant No. 2139899.

\bibliographystyle{unsrtnat}
\bibliography{references}

\end{document}